%% file: main.tex
\documentclass[iicol]{sn-jnl}
\usepackage[numbers]{natbib} %
\usepackage{graphicx}%
\usepackage{multirow}%
\usepackage{amsmath,amssymb,amsfonts}%
\usepackage{amsthm}%
\usepackage{mathrsfs}%
\usepackage[title]{appendix}%
\usepackage{xcolor}%
\usepackage{textcomp}%
\usepackage{manyfoot}%
\usepackage{booktabs}%
\usepackage{algorithm}%
\usepackage{algorithmicx}%
\usepackage{algpseudocode}%
\usepackage{listings}%
\usepackage{float}%
\usepackage{subfigure}

\def\ie{\textit{i.e.}}

\def\x{\mathbf{x}}

\def\z{\mathbf{z}}

\theoremstyle{thmstyleone}%
\theoremstyle{thmstyletwo}%

\theoremstyle{thmstylethree}%
\newcommand{\yf}[1]{{\color{black} #1}}

\begin{document}

\title[Article Title]{Mastering Regional 3DGS: Locating, Initializing, and Editing \\ with Diverse 2D Priors}

\author[1]{\fnm{Lanqing} \sur{Guo}}\email{lanqing.guo@austin.utexas.edu}

\author[2]{\fnm{Yufei} \sur{Wang}}\email{im.wangyufei@gmail.com}
\author[1]{\fnm{Hezhen} \sur{Hu}}\email{hezhen.hu@austin.utexas.edu}
\author[1]{\fnm{Yan} \sur{Zheng}}\email{yanzheng@utexas.edu}
\author[3]{\fnm{Yeying} \sur{Jin}}\email{jinyeying@u.nus.edu}
\author[4]{\fnm{Siyu} \sur{Huang}}\email{siyuh@clemson.edu}
\author[1]{\fnm{Zhangyang} \sur{Wang}}\email{atlaswang@utexas.edu}


\affil[1]{\orgname{The University of Texas at Austin},  \orgaddress{\city{Austin}, \country{USA}}}
\affil[2]{\orgname{Snap Research}, \orgaddress{\city{NYC}, \country{USA}}}
\affil[3]{\orgname{Tencent},  \orgaddress{\country{Singapore}}}
\affil[4]{\orgname{Clemson University},  \orgaddress{\city{Clemson}, \country{USA}}}

\abstract{Many 3D scene editing tasks focus on modifying local regions rather than the entire scene, except for some global applications like style transfer, and in the context of 3D Gaussian Splatting (3DGS), where scenes are represented by a series of Gaussians, this structure allows for precise regional edits, offering enhanced control over specific areas of the scene; however, the challenge lies in the fact that 3D semantic parsing often underperforms compared to its 2D counterpart, making targeted manipulations within 3D spaces more difficult and limiting the fidelity of edits, which we address by leveraging 2D diffusion editing to accurately identify modification regions in each view, followed by inverse rendering for 3D localization, then refining the frontal view and initializing a coarse 3DGS with consistent views and approximate shapes derived from depth maps predicted by a 2D foundation model, thereby supporting an iterative, view-consistent editing process that gradually enhances structural details and textures to ensure coherence across perspectives. Experiments demonstrate that our method achieves state-of-the-art performance while delivering up to a $4\times$ speedup, providing a more efficient and effective approach to 3D scene local editing.}

\keywords{3D Gaussian Splatting, Image Generation, 3D Editing}
\maketitle

\input{sec/1_intro}
\input{sec/2_method}

\input{sec/3_exper}

\input{sec/4_concl}

\section{Data Availability Statement}
The datasets analyzed during the current study are available in the repository at the following link: \url{https://instruct-gs2gs.github.io/}.
\section{Declarations}
\noindent\textbf{Conflict of interest.} The authors declare that they have no Conflict of
interest.
\backmatter

\bibliography{sn-bibliography}

\end{document}

%% file: sec/1_intro.tex
\section{Introduction}
\label{sec:intro}

\begin{figure*}[t!]
    \begin{center}
        \centering
        \maketitle
        \includegraphics[width=1.0\textwidth]{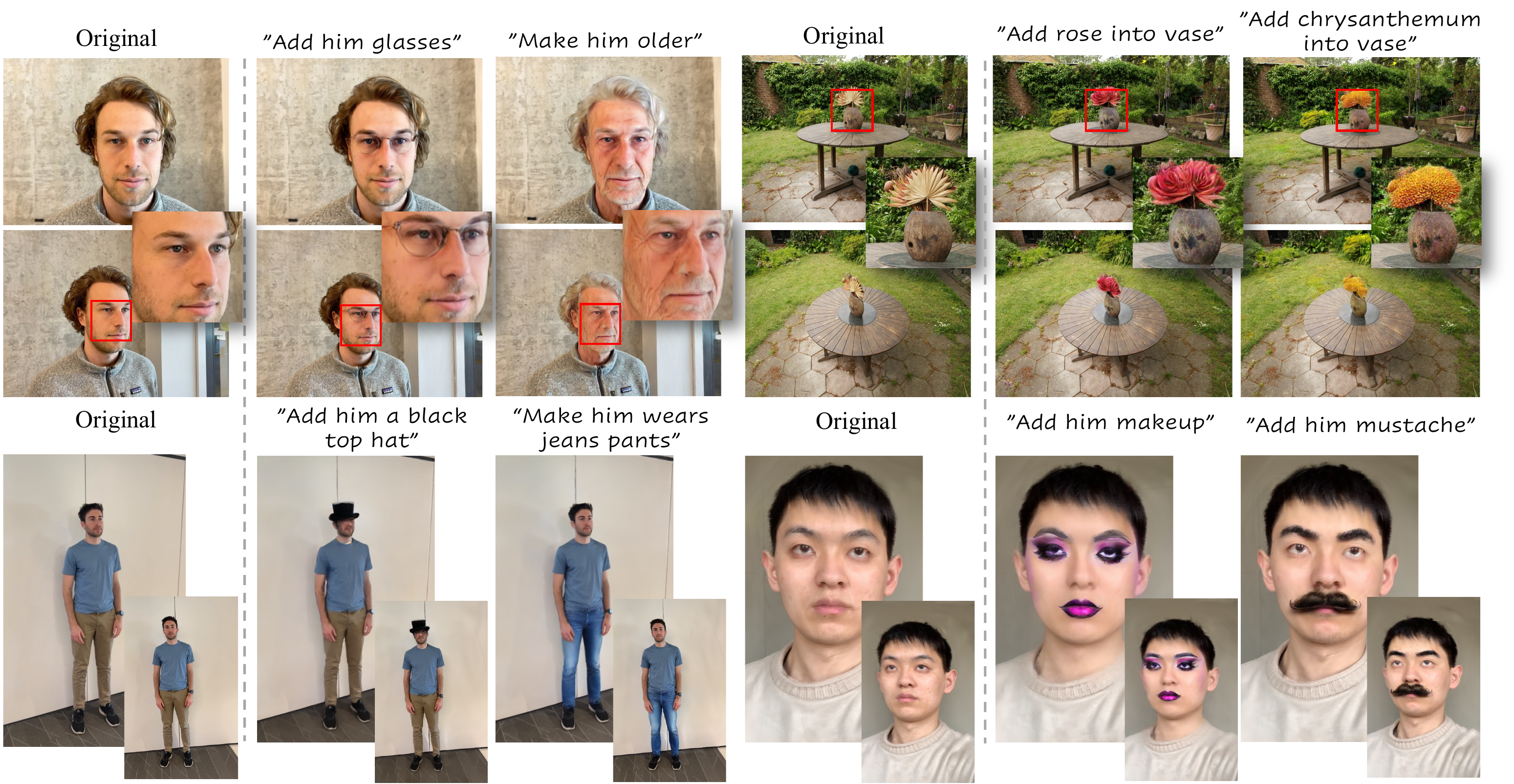}
        \caption{Comprehensive results demonstrate our method’s effectiveness in diverse localized editing tasks, including modifications to human bodies, faces, and scenes. The leftmost column displays the original view, while the two columns on the right show rendered views of the edited 3DGS generated by our approach using the corresponding text descriptions. Please zoom in for a clearer visualization.}
        \label{fig:illustration}
    \end{center}
\end{figure*}


In many 3D scene editing applications~\cite{zuo2025fmgs}, the focus is increasingly on modifying specific aspects, such as avatar attribute editing~\cite{ho2023custom,zhu2024InstructHumans}, adding or removing objects~\cite{ruiz2024magic,anciukevivcius2023renderdiffusion,weber2024nerfiller}, and similar localized adjustments. Unlike global modifications like style transfer~\cite{chen2024upst,jain2024stylesplat}, which aim to achieve a consistent look or feel across an entire scene, localized edits allow for targeted changes that enhance specific elements without altering the overall scene context, which are also aligned with various nuanced requirements of tasks like object manipulation, detailed feature enhancement, or region-specific alterations.
These targeted modifications are particularly valuable in applications requiring precision, where fine-grained control over specific parts of a scene is essential for achieving high visual fidelity. 
One emerging 3D modeling approach in this space, 3D Gaussian Splatting (3DGS)~\cite{kerbl3Dgaussians}, represents scenes as a series of Gaussian points, which allows for accurate regional adjustments and enhanced control over discrete areas of the 3D environment.
This representation is highly flexible, enabling it to facilitate localized scene modifications with greater accuracy and precision.

However, despite the advantages offered by 3DGS in terms of localization, there remain considerable challenges in achieving high-quality, consistent edits across all views in a 3D space. Many current methods for 3D editing struggle with two primary issues: time efficiency and detail fidelity. Existing approaches~\cite{chen2024gaussianeditor,haque2023instruct} often rely on time consuming diffusion model editing rendered frames and iterative optimization across multiple views, resulting in extended processing times that hinder real-time application.
Additionally, these methods frequently exhibit detail loss, particularly in intricate structures or textures. 
The underperformance of 3D semantic parsing relative to 2D parsing further complicates the editing process, as it becomes difficult to reliably identify and target specific regions within the 3D space without sacrificing the quality of edits. 
As a result, achieving high fidelity without compromising edit quality remains difficult, especially in intricate scenes where detail is critical to preserving visual coherence and realism.

To address these issues, we propose a novel approach that combines 2D diffusion editing with a re-rendering process to achieve efficient, high-fidelity localization within 3D environments. Our method begins with 2D diffusion-based region identification, enabling precise targeting of edit regions in each view without the need for time-consuming 3D parsing steps. Once these regions are identified in 2D, we re-render the scene and use Gaussian point expansion to locate the corresponding areas within the 3D space. This approach not only accelerates the editing process but also circumvents the limitations of conventional 3D parsing by leveraging the higher accuracy of 2D parsing models. Following this step, we initialize a coarse 3D Gaussian Splatting representation using consistent views and approximate shapes derived from depth maps predicted by a 2D foundation model. This initialization serves as a strong foundation for an iterative, view-consistent editing process that progressively refines both structural details and textures, ensuring coherence and detail preservation across all perspectives.
Experimental results demonstrate that our method achieves state-of-the-art performance, providing substantial improvements in both accuracy and speed.

Our main contributions are concluded as follows:
\begin{itemize}
\item We propose a new 3DGS editing framework that separates the process into sequential localizing, initializing, and editing stages, ensuring explicit Gaussian points localization and precise modification.


    \item Within the located Gaussians, we initialize a coarse 3DGS with approximate shapes based on the depth map predicted by the 2D foundation model. Then we refine the structural details and textures of sequential views by conditional sequential diffusion editing.

    \item Experimental results demonstrate that our 3DGS local editing method achieves both higher efficiency and better structural details compared to existing methods, achieving up to $4\times$ speedup comparing to 3DGS-based editing.

\end{itemize}

\section{Related Work}

\subsection{Text-based Image Editing}

Given the limited availability of large-scale datasets specifically designed for training 3D editors from scratch, most existing methods circumvent this challenge by adapting techniques originally developed for 2D image editing. In particular, the field of text-to-image editing—driven by the success of diffusion models—has witnessed rapid progress in recent years~\cite{quan2024deep,wang2025training,zheng2024oscillation,xiao2024fastcomposer,wang2025lavie,wang2024exploiting,zhang2024show}. Early methods in this line of work typically apply noise to the source image and then re-generate it based on a new prompt, aiming to achieve the desired edit through conditional denoising. However, such approaches often fall short in preserving data fidelity; the edited results frequently suffer from layout distortions or the loss of critical structural and semantic content.
To address these shortcomings, subsequent methods have explored techniques to better preserve the spatial and semantic integrity of the original image. Some approaches attempt to align attention maps during generation~\cite{hertz2022prompt,mokady2023null}, ensuring that important regions remain stable throughout the editing process. Others modify internal feature representations~\cite{parmar2023zero,tumanyan2022splicing}, effectively steering the generation path closer to the original content. These methods have also inspired instruction-based editing pipelines, such as InstructPix2Pix~\cite{brooks2023instructpix2pix}, which leverages large-scale synthetic data conditioned on both images and editing instructions to train a dedicated diffusion model. Building on this paradigm, further efforts have refined InstructPix2Pix via human-annotated supervision to improve edit controllability and consistency.
In a different direction, another class of methods directly fine-tunes the diffusion model on the source image itself~\cite{kawar2023imagic,valevski2023pix2pixzero}, enabling the model to better internalize the structure and appearance of the specific image. This approach significantly improves content alignment, ensuring that both high-level semantics and fine-grained visual details are faithfully preserved throughout the editing process. Together, these advancements form the foundation for recent progress in generative editing, which is increasingly being adapted and extended to the more complex domain of 3D content manipulation.

\subsection{Text-driven 3D Editing}

Recently, image editing techniques have expanded into 3D scene consistent editing. NeRF~\cite{mildenhall2020nerf} and 3DGS~\cite{kerbl20233d} are among the most popular 3D models for neural novel view synthesis: NeRF encodes scene geometry and color implicitly through a Multi-Layer Perceptron (MLP), while 3DGS represents the scene explicitly with Gaussian ellipse point clouds. IN2N~\cite{haque2023instruct} pioneered NeRF-based 3D editing with text instructions, utilizing 2D image editing tool, \ie, IP2P~\cite{brooks2023instructpix2pix}, on rendered views and iteratively updating the 3D scene until convergence. Subsequently, ViCA-NeRF~\cite{dong2024vica} introduced cross-view blending to improve view consistency, and DreamEditor~\cite{zhuang2023dreameditor} incorporated SDS loss and DreamBooth~\cite{ruiz2023dreambooth} to address object consistency issues. However, NeRF-based editing faces challenges with slow processing speeds and limited control over specific scene regions.
With advances in 3DGS, new 3DGS-based editing approaches have emerged. For example, GaussianEditor~\cite{chen2024gaussianeditor} leverages 2D IP2P as a prior and introduces semantic tracing to track the editing target throughout training. Later methods, such as DGE~\cite{chen2024dge} and VcEdit~\cite{wang2025view}, propose techniques like the episolar constraint and Cross-attention Consistency Modules to enhance view consistency in edits. Nonetheless, semantic-level consistency constraints still suffer from low-level structural inconsistencies, limiting the fidelity of edits. Besides, the editing process still face challenge for practical applications.
In this paper, we delve into the realm of 3DGS to explore a highly efficient approach for 3D local editing. We propose a coarse-to-fine pipeline that initializes 3DGS using a predicted coarse Gaussian points by predicted depth maps and refines structural details through conditional sequential 2D view editing.

\subsection{Local Editing}


Local editing refers to modifications confined to specific regions within 2D or 3D scenes, encompassing a wide range of tasks such as object manipulation, attribute adjustment~\cite{zhu2016generative}, spatial transformation~\cite{jaderberg2015spatial}, and inpainting~\cite{lugmayr2022repaint}, among others~\cite{shuai2024survey}. For certain explicit local editing tasks, a binary mask indicating the region of interest is often provided as an auxiliary input to guide the model~\cite{isola2017image}, ensuring edits are spatially constrained. To further improve precision and avoid unintended alterations, recent works~\cite{meng2021sdedit,lugmayr2022repaint} have introduced diffusion-based editing frameworks that effectively control modification strength and localization. Notably, \cite{brooks2023grounding} extends this paradigm to support multi-region editing by applying a DDS loss to designated areas of the image, thus enabling localized pixel optimization while preserving surrounding content.
In contrast to the substantial progress in 2D, local editing in 3D scenes—particularly at the semantic or attribute level—remains relatively underexplored due to challenges in region localization, multi-view consistency, and the lack of fine-grained 3D supervision. Some early efforts have made progress in this direction: for instance, \cite{xiao2024localized} explores object-level insertion and removal operations to coarsely infer editable regions in 3D space, while \cite{mirzaei2025watch} proposes using relevance maps from 2D projections as editing masks, which, however, may lead to view inconsistency and suboptimal control in the 3D domain.
In this paper, we propose to extend the concept of localization from 2D to 3D by operating on semantically meaningful regions tied to specific attributes, and we build upon the 3D Gaussian Splatting (3DGS) representation to enable precise positioning and manipulation of Gaussians within the identified regions, supporting coherent and fine-grained local edits in 3D scenes.

%% file: sec/2_method.tex
\section{Preliminary}

\begin{figure*}[t!]
    \centering
    \includegraphics[width=1.\textwidth]{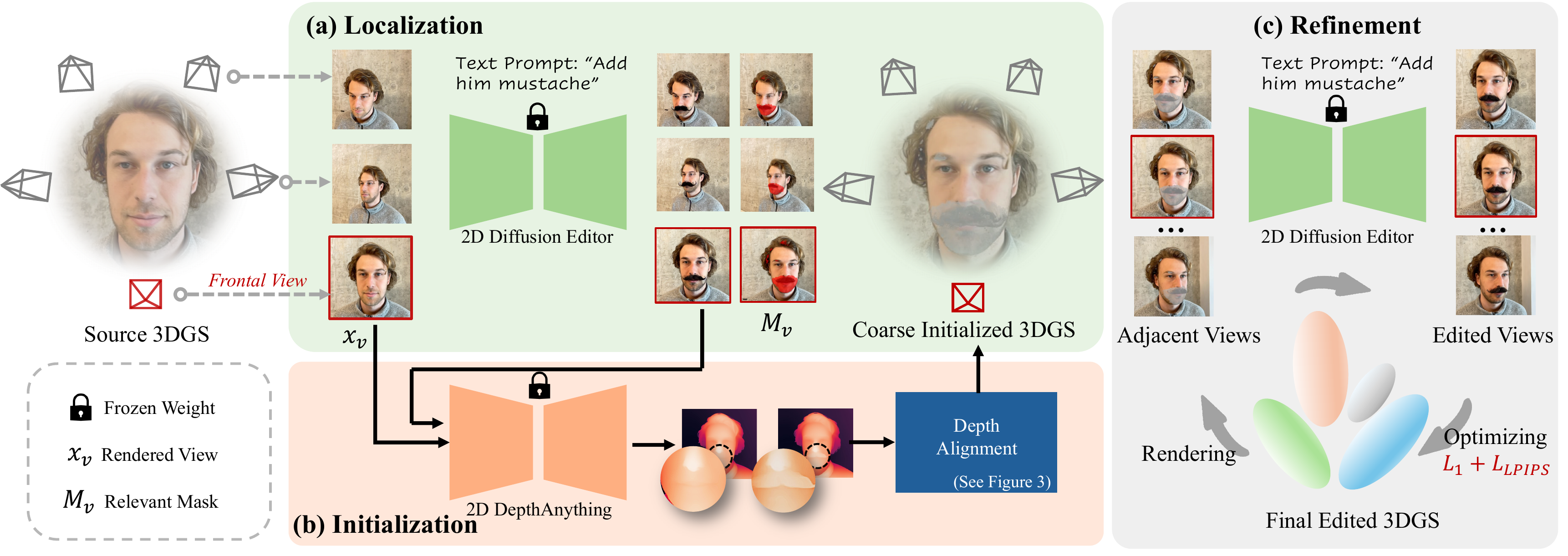}
    \caption{
    The overall framework of the proposed method that enables efficient fine-grained editing to 3DGS scenes. \textbf{Localization:} 
    Given the text prompt, we first isolate relevant regions for each view by evaluating noise differences with and without text guidance (Eq.\ref{eq:mask2d}). The resulting 2D masks are inverse-rendered to 3D~\cite{chen2024gaussianeditor,wang2025view} to obtain a view-consistent 3D mask, which has better quality in the side views that originally IP2P fails. 
    \textbf{Initialization:} Local editing not only affects the existing points in the source 3DGS, but it may also require newly added Gaussian points, \textit{e.g.}, added mustache. To address this, we propose leveraging the frontal view's depth prior from a pretrained monocular depth estimator.
    Then, we add new Gaussian points by mapping the edited pixels in the aforementioned 2D mask to 3D space using the monocular depth. \textbf{Refinement:} We refine views adjacent to the frontal view conditioned on the coarsely edited scene, where the 3DGS scene is iteratively finetuned using refined views. After several finetuning iterations, a new frontal view is selected, and the adjacent refinement and optimization steps are repeated. (We visualize mask $M_v$ as the mixed picture with corresponding image for visualization.)
    }
    \label{fig:method}
\end{figure*}

\subsection{3D Gaussian Splatting}
3D Gaussian Splatting (3DGS) models scenes using anisotropic 3D neural Gaussians for efficient rendering via tile-based rasterization. Each Gaussian \( G \) is characterized by its mean \( \mu \in \mathbb{R}^3 \), covariance \( \Sigma \), color \( c \in \mathbb{R}^3 \), and opacity \( \alpha \in \mathbb{R} \). To optimize, the covariance \( \Sigma \) is decomposed as \( \Sigma = R S S^T R^T \), with scaling \( S \in \mathbb{R}^3 \) and rotation \( R \in \mathbb{R}^{3 \times 3} \). A Gaussian centered at \(\mu\) is given by \( G(x) = \exp \left( -\frac{1}{2} (x - \mu)^T \Sigma^{-1} (x - \mu) \right) \), where \( x \) is the 3D coordinates. The 3DGS \( \mathcal{G} \) can be efficiently projected to 2D space, where the color $C(x')$ depth \( D(x') \) for each pixel \( x' \) can be obtained via point-based volumetric rendering:
\begin{align}
    {C}(x') &= \sum_{i \in N} c_i \sigma_i \prod_{j=1}^{i-1} (1 - \sigma_j),\\
        {D}(x') &= \sum_{i \in N} d_i \sigma_i \prod_{j=1}^{i-1} (1 - \sigma_j), \quad \sigma_i = \alpha_i G(x')
        \label{eq:depth_gs}
\end{align}
where $N$ is the ordered set of Gaussians overlapping the pixel $x'$, and $d_i$ is the distance between 3D Gaussian center $\mu_i$ and camera center.

\subsection{Diffusion-based Editing}

A diffusion-based image editing method samples the target image $\x'$ from a conditional distribution $p(\cdot|\x, c)$, where the condition $c$ can be a textual prompt for text-based editing and $\x$ is the source image.
We adopted \yf{IP2P~\cite{brooks2023instructpix2pix} as the 2D prior for view editing, which is built upon Stable Diffusion (SD)~\cite{podell2023sdxl} and utilizes the classifier-free guidance~\cite{ho2022classifier} during inference.}
\yf{Specifically, the encoder $\mathcal{E}$ first converts image into latent code $\z'_0 = \mathcal{E}(\x_0)$, and the sampling with the guidance of $\mathbf{c}$ is defined as follows
\begin{equation}
\tilde{\epsilon}_{\theta}(\mathbf{z}_{\lambda}, \z'_0, \mathbf{c}) = (1 + w) \epsilon_{\theta}(\mathbf{z}_{\lambda}, \z'_0, \mathbf{c}) - w \epsilon_{\theta}(\mathbf{z}_{\lambda}, \z'_0),
\end{equation}
where $\epsilon_{\theta}(\mathbf{z}_{\lambda}, \z'_0, \mathbf{c})$ and $\epsilon_{\theta}(\mathbf{z}_{\lambda}, \z'_0)$ are the predicted noise w/ and w/o the condition guidance, and $w$ is the weight of the guidance. The difference to~\cite{ho2022classifier} is that IP2P additionally conditioned on the latent of the input image $\z_0'$ to support image editting.}
During the forward diffusion process, noise is gradually added to the latent code:
\begin{align}
    q(\z'_t|\z'_0) = \mathcal{N}(\z'_t; \sqrt{\bar{\alpha}_t} \z'_0, (1-\bar{\alpha}_t)\mathbf{I})\;.
    \label{eq:addnoise}
\end{align}
where $\bar{\alpha}_t$ controls the noise level at timestep $t$.

\section{Methodology}
In this section, we present the overall pipeline of our proposed method. First, we perform 3D localization to identify Gaussians requiring edits based on accurate 2D location information (Section~\ref{sec:locate}). Next, we initialize a coarse 3D Gaussian points within the identified 3D regions using predicted depth (Section~\ref{sec:init}). Finally, we apply conditional sequential view refinement to enhance texture, color, and structural details (Section~\ref{sec:refine}). The complete framework is illustrated in Figure~\ref{fig:method}.


\subsection{From 2D to 3D Localization}\label{sec:locate}

The essence of the editing task can be decomposed into locating the modification regions based on user requirements and then adjusting these regions while preserving information from the original image. 
In the first step, we explicitly localize the target Gaussian points in 3D space, identifying the specific points to be modified.


Inspired by previous work~\cite{mirzaei2025watch}, given several rendered views $\mathbf{X} = (\x_v)_{v=1}^V$ from the source 3DGS $\mathcal{G}$ and a textual prompt $c$ describing the desired edits, we can perform IP2P to isolate regions relevant to the prompt. We achieve the 2D edited areas $M_v$ for each view by calculating the difference between the noise predicted by IP2P's UNet, $\epsilon_\theta$, when conditioned on both the image and text, and the noise predicted when conditioned only on the image with an empty text prompt as follows
\yf{
\begin{align}
    {M}_{v} = \mathcal{F}( \left\|\epsilon_\theta(\z_{v,\tau}, \x_v, \mathbf{c}) - \epsilon_\theta(\z_{v,\tau}, \x_v, \emptyset) \right\|_{c}, \gamma),
\label{eq:mask2d}
\end{align}
where $\tau$ is a hyper-parameter determines the timestep that we use, $\z_{v,\tau} = \sqrt{\alpha_\tau} \, \mathcal{E}(\x_v) + \sqrt{1 - \alpha_\tau} \, \epsilon$ is the noised latent code in timestep $\tau$, and $\left\|\cdot\right\|_{c}$ is the mean absolute value in the channel dimension. We obtain the binary mask through $\mathcal{F}$ by applying a threshold \(\gamma \in [0,1]\), followed by Gaussian filtering to reduce isolated regions and noise.}


\yf{After that, we inverse render $\{M_v\}_{v=1}^V$ to 3D space~\cite{wang2024gaussianeditor,wang2025view} to achieve the sparse 3D mask $\mathcal{M}$ with the number of Gaussian points in the source 3DGS $\mathcal{G}$.} 

\subsection{Incremental Initialization with Predicted Depth}\label{sec:init}
\begin{figure}[t!]
    \centering
    \includegraphics[width=1.\linewidth]{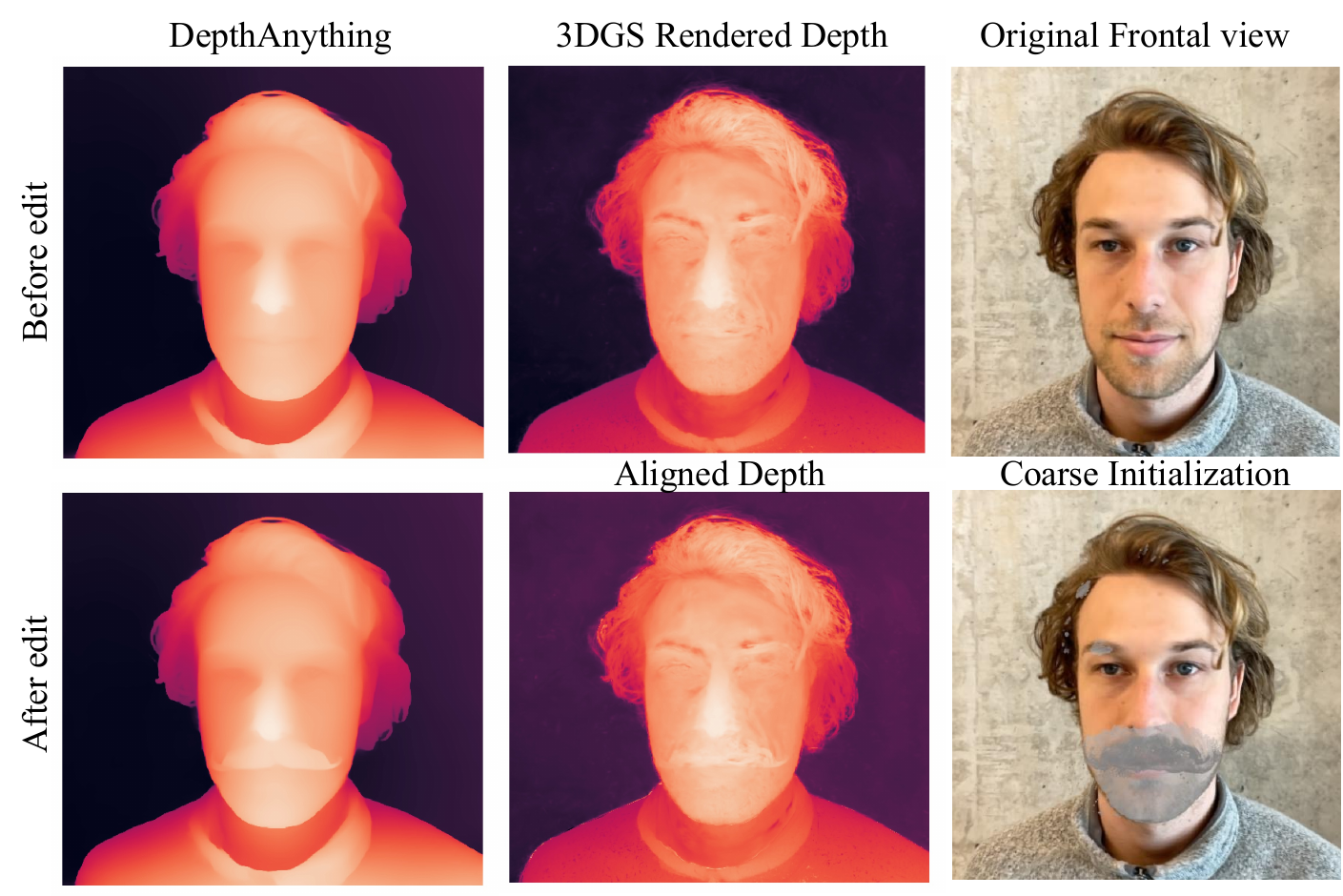}
    \caption{Illustration of the local initialization process using depth maps predicted by DepthAnything~\cite{yang2024depth,yang2024depth1}. 
    According to the predicted depth maps by DepthAnything on before and after edit 2D view images, we convert the 3DGS rendered depth map into aligned depth, adjusted based on the residual differences between the before and after DepthAnything predictions. This yields a regional initialized Gaussian points and its corresponding coarse rendered edit view.
    }
    \label{fig:depth}
\end{figure}

Although 2D diffusion generation/editing models achieve more promising results than 3D models, they \yf{usually only} perform well on specific views, such as the frontal view, due to the higher amount of training data for these perspectives. \yf{Therefore, we propose to randomly select a few views and start by editing the view $v_{\text{first}}=\arg \max_{v} \sum_{i,j}{M}_{v,i,j}$, which has the largest correlation with the text prompt}.
To provide better initialization of the the Gaussian points, which was demonstrated to be crucial especially for sparse view settings~\cite{zhu2025fsgs}, we propose to utilize the depth prior from Depth Anything model~\cite{yang2024depth,yang2024depth1}. Specifically, since the output of the depth anything is the disparity instead of the real depth as shown in Figure~\ref{fig:depth}, we need to calibrate it first before usage. Inspired by \cite{kerbl2024hierarchical}, we propose to first utilize the depth rendered by the unedited Gaussian points using Eq. \ref{eq:depth_gs} to obtain the unknown scale $a$ and bias $b$ as follows which we found has better robustness than directly using the linear regression
\begin{equation}
    a = \frac{s(\delta_{3DGS})}{s(\delta_{mono})},
    b= f(\delta_{3DGS})-a\cdot f(\delta_{mono}),
\end{equation}
where $\delta_{mono}$ and $\delta_{3DGS}$ are the disparities from depth anything and 3DGS, respectively, \textit{i.e.}, $\mathbf{d}=\frac{1}{\delta}$, $f(\delta)=\text{median}(\delta)$ and $s(\delta)=\frac{1}{M}\sum_{i}|\delta_i-f(\delta)|$. We could obtain the depth after calibration from monocular depth estimator as $\mathbf{d}_{mono}=\frac{1}{a\delta_{mono}+b}$. After that, we can obtain the Gaussian points to be added as follows 
\begin{equation}
    \Delta \mathcal{G} = \mathcal{M} \odot P[\frac{\mathbf{d}_{mono}(\x_{edited})}{\mathbf{d}_{mono}(\x_{unedit})} \mathbf{d}_{3dgs}, \x ^{edit}]
\label{eq:depth_trans}
\end{equation}
where $P$ is the mapping from 2D pixels to 3D points given the predicted depth and image, and $\mathbf{d}_{mono}(\x)$ is the calibrated monocular depth from image $\x$. The motivation behind 
Eq.~\ref{eq:depth_trans} is that it uses the same estimator to identify the relative depth variation, enhancing the robustness of the results. The added Gaussian points $\Delta \mathcal{G}$ are added to the source 3DGS to provide better initialization of areas with depth change.


\begin{table*}[t]
    \centering
        \caption{When comparing different editing methods, Gaussian Splatting-based techniques are considerably quicker than NeRF-based approaches. Our method achieves comparable performance, particularly in preserving data fidelity, while offering the fastest editing speed.}
    \begin{tabular}{l|c|cccc}
    \toprule
    Method & 3D Model   & CTIDS$\uparrow$ & IIS$\uparrow$ & FID$\downarrow$ & Editing Time \\
    \midrule
        IN2N~\cite{haque2023instruct} & NeRF& \textbf{0.2128} &\textbf{0.7591} & {150.80}& $\sim 51$min \\
         ViCA-NeRF~\cite{dong2024vica} & NeRF & 0.2103 & 0.7205 & 161.35 & $\sim 28$min\\
         WatchYourSteps~\cite{mirzaei2025watch} & NeRF & {0.2117} & 0.7368 & \textbf{129.68} & $\sim 51$min \\
         \midrule
        GaussianEditor~\cite{chen2024gaussianeditor} & 3DGS & 0.2043  & \underline{0.8212}& 70.14 &$\sim 7$min \\
        DGE~\cite{chen2024dge} & 3DGS & \textbf{0.2057} & 0.8183 &\underline{68.60} & $\sim 4$min\\
        Gaussctrl~\cite{gaussctrl2024} & 3DGS & 0.1882 & 0.7862& 132.76 & $\sim 9$min\\
        Ours & 3DGS & \underline{0.2053}& \textbf{0.8308} & \textbf{65.96} & $\sim 2$min\\
        \bottomrule
    \end{tabular}
    \label{tab:my_label}
\end{table*}

\subsection{Conditional Sequential View Refinement}\label{sec:refine}
After the initialization, for viewpoints near the frontal view, the 3D shape already has a substantial initialization and only requires refinement of texture and color. Additionally, the context information provided by the text-irrelevant regions $(1-\mathcal{M})$ makes these viewpoints easier to edit.
To this end, we employ a conditional sequential view refinement approach to enhance the quality of each frame progressively.
Specifically, we select $m$ adjacent views $\{\x_{j_k}\}_{k=1}^m$ of the frontal view $\x_{v_{\text{first}}}$: 
\begin{align}
    \{\x_{j_k}\}_{k=1}^m = \operatorname{arg \, top}_m \left( \|\mathbf{p}_{v_{\text{first}}} - \mathbf{p}_i\| \mid v_{\text{first}} \neq i \right),
\end{align}
where $\|\mathbf{p}_{v_{\text{first}}} - \mathbf{p}_i\| $ represents the Euclidean distance between the camera positions of views $\x_{v_{\text{first}}}$ and $\x_i$.

We then apply IP2P to refine each frame in \(\{\x_{j_k}\}_{k=1}^m\), enhancing both the visual fidelity and temporal coherence across frames. This refinement is achieved by conditioning each frame on both the previously initialized edits \(\tilde{\x}_{j_k}\) and the original unedited frame \(\x_{j_k}\). 
We follow previous methods~\cite{haque2023instruct} employing IP2P to sample the final refined image \(\x'_{j_k}\) from the distribution \(p(\x'_{j_k} | \x_{j_k}, \tilde{\x}_{j_k}, \mathbf{c})\).
We add noise to the latent code of coarse views, \(\tilde{\x}_{j_k}\), as defined in Eq.~(\ref{eq:addnoise}), generating \(\tilde{\z}_{j_k, t}\). During IP2P sampling, denoising begins from \(\tilde{\z}_{j_k, t}\) at timestep \(t\).
This approach ensures that each frame not only retains the intended edits but also maintains consistency and smooth transitions with adjacent frames, leading to a cohesive and high-quality visual output.
Those edited frames will be fed to optimize the 3DGS scene as follows:
\begin{align}
    \mathcal{G}^{\text{edit}} = \arg\min_{\mathcal{G}} \sum_{v \in V} \mathcal{L}_{\text{edit}} (\mathcal{R}(\mathcal{G}, v), \x'_v)
\end{align}
where $\mathcal{L}_{\text{edit}}$ is employed to reconstruct a 3D scene from corresponding edited views, incorporating LPIPS loss~\cite{zhang2018unreasonable} and L1 loss as used in prior work~\cite{chen2024dge}. Here we still optimize the whole Gaussian points to avoid the inaccurate initialization and large shape editing cases.

After several iterations of finetuning the 3DGS, we designate a new frontal view \yf{from $\{\x'_v\}_{v\in V}$} as the updated reference and reselect $m$ adjacent views to initiate a fresh round of conditional editing. The \yf{workflow for a single round} can be summarized as follows: randomly select a frontal view $\rightarrow$ select $m$ adjacent views of the chosen frontal view $\rightarrow$ apply 2D conditional editing $\rightarrow$ perform 3DGS finetuning. Each cycle progressively improves coherence across perspectives, ensuring consistency and alignment with the updated reference viewpoint. To improve efficiency, we progressively reduce the IP2P sampling starting timestep $t$ in experiments. We perform two or three such cycles in experiments. The detailed hyper-parameters can be found in supplementary materials.

%% file: sec/3_exper.tex
\section{Experiments}

\begin{figure*}[t!]
    \centering
    \includegraphics[width=1.\textwidth]{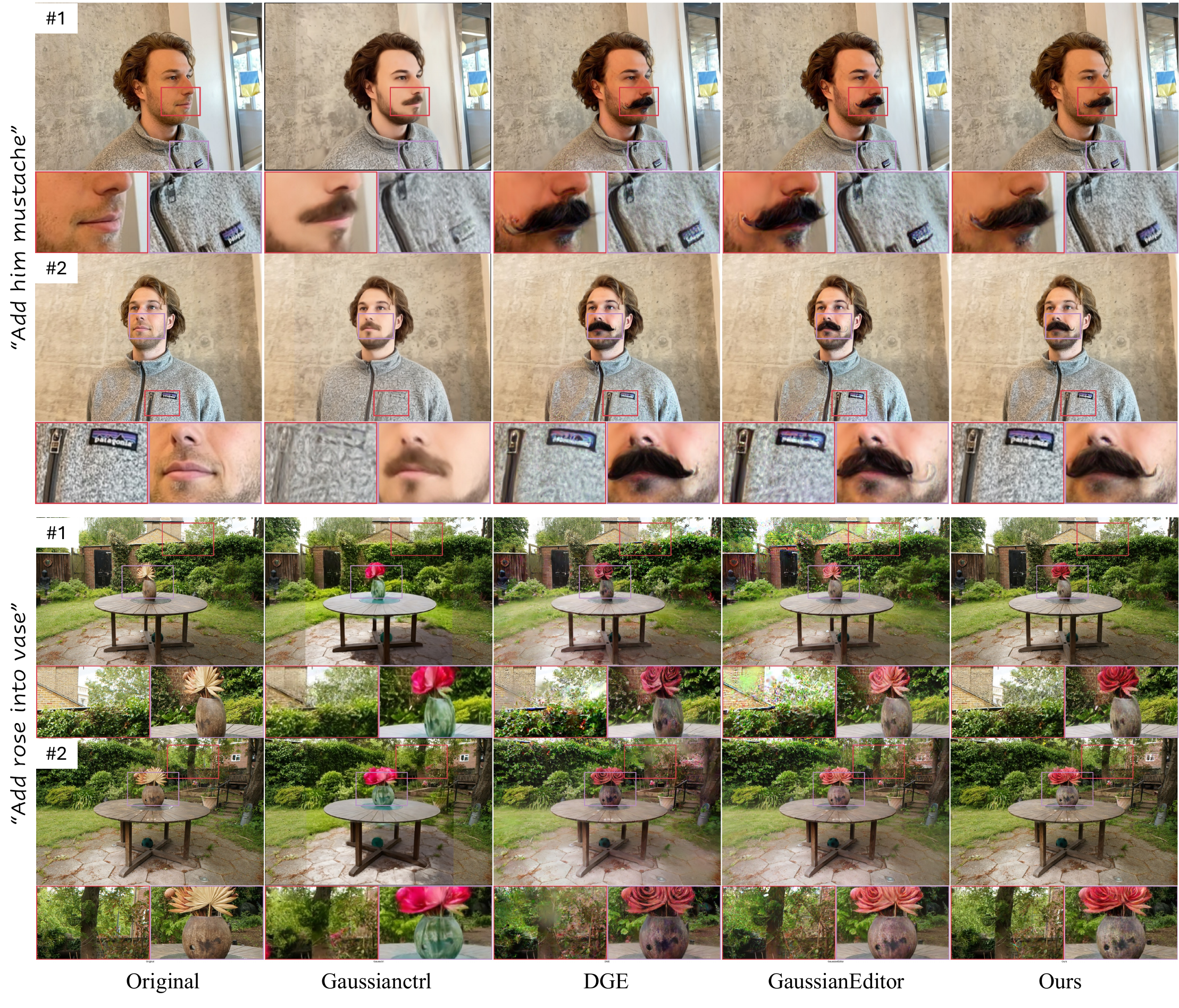}
    \caption{Qualitative comparison with three 3DGS-based competing methods: The leftmost column demonstrate the original views, while the right four columns show the rendering view of edited 3DGS by Gaussianctrl~\cite{gaussctrl2024}, DGE~\cite{chen2024dge}, GaussianEditor~\cite{chen2024gaussianeditor}, and our method, respectively. Our method effectively preserves fidelity and structural details while delivering artifact-free and view-consistent edited results.
}
    \label{fig:res}
\end{figure*}

\begin{figure*}[t!]
    \centering
    \subfigure[Examples on ``person'' scene from IN2N~\cite{haque2023instruct}.]{
        \includegraphics[height=10cm]{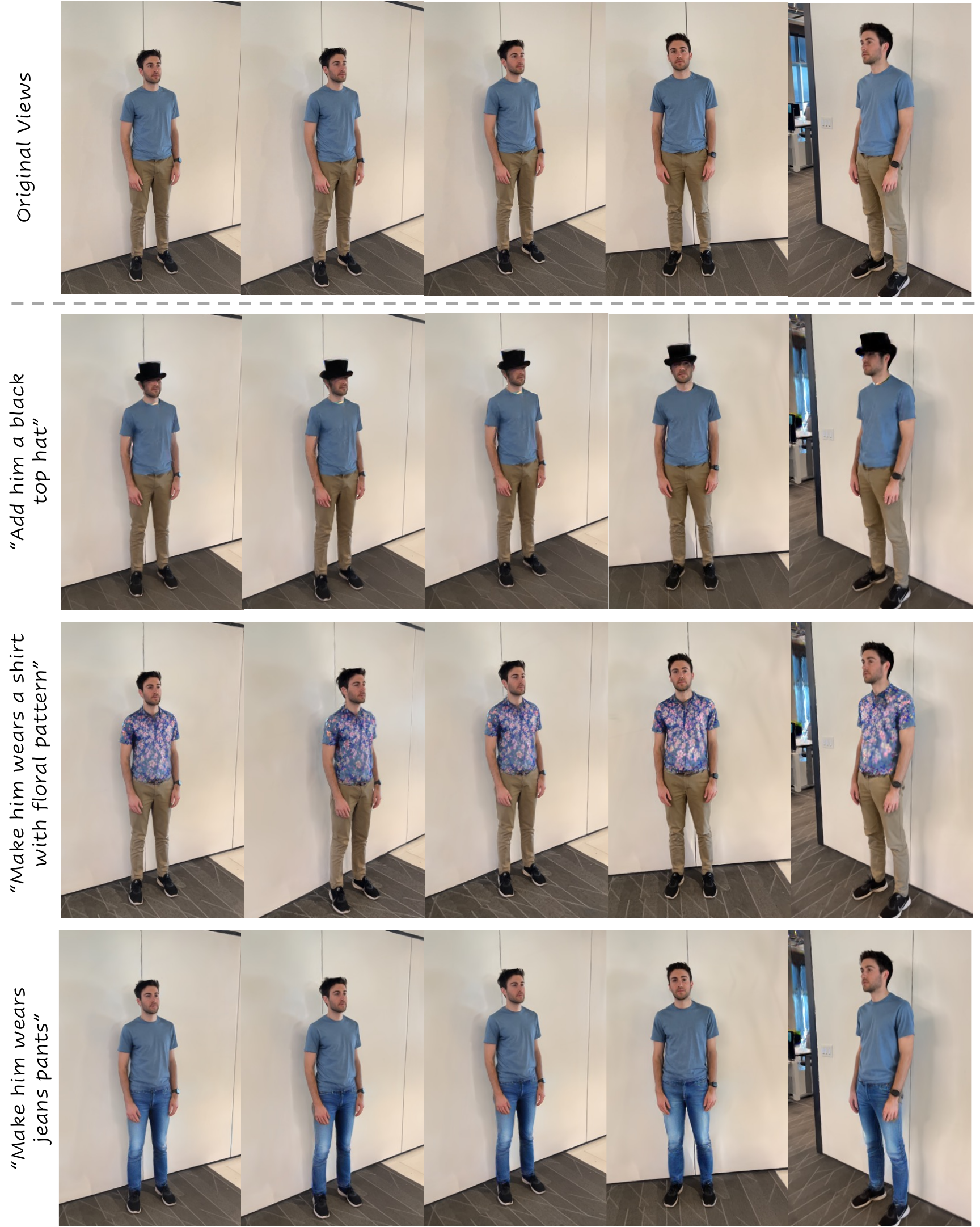}
        \label{fig:res_person}
    }
    \hfill
    \subfigure[Examples on ``fangzhou'' scene from Nerf-Art~\cite{wang2023nerf}.]{
        \includegraphics[height=10cm]{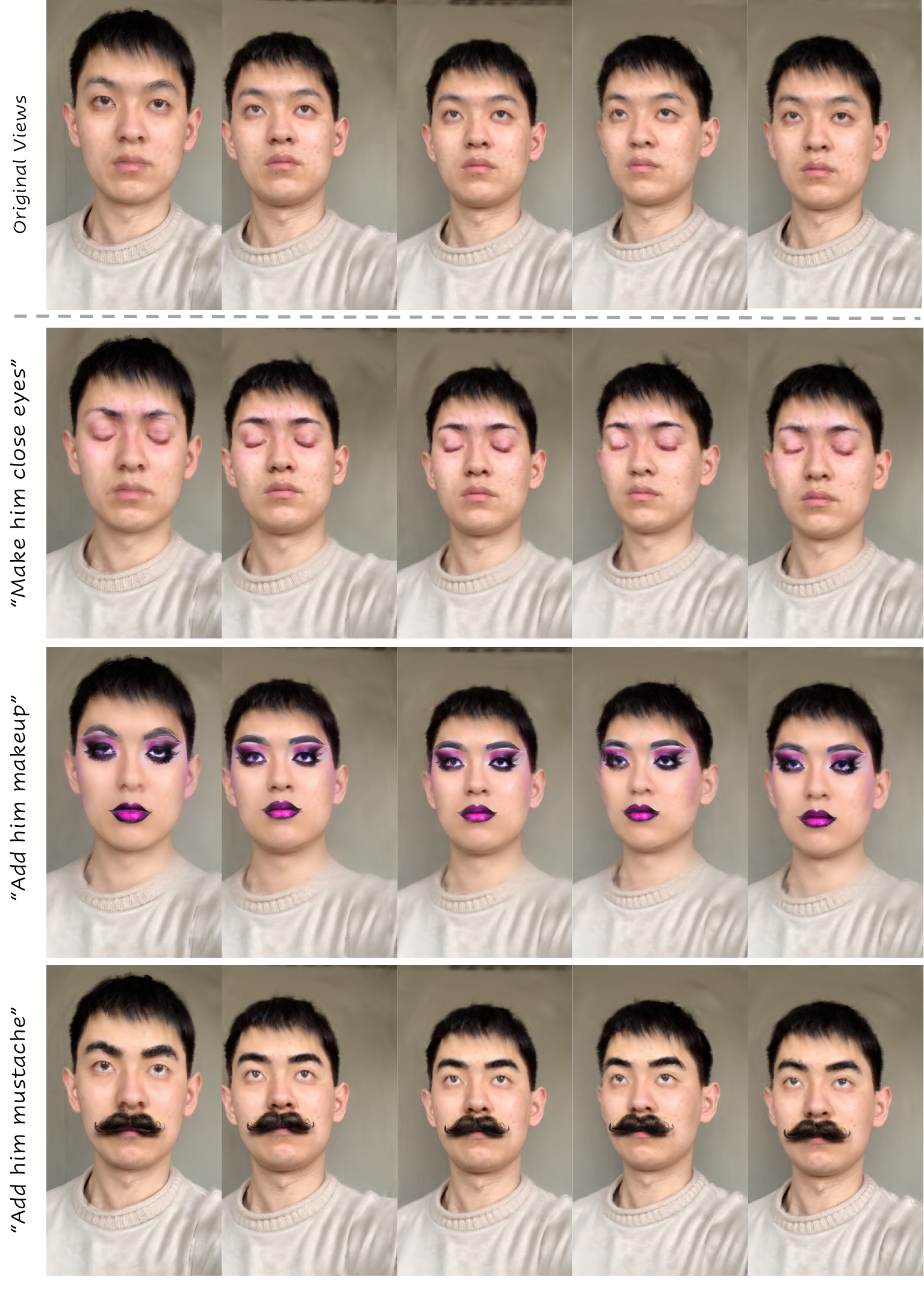}
        \label{fig:res_fangzhou}
    }
    \caption{Illustrative visual examples showcasing the application of our proposed method for local editing on 3D scenes, demonstrating its effectiveness in handling fine-grained modifications.}
    \label{fig:res_combined}
\end{figure*}

\begin{figure*}[t!]
    \centering
    \includegraphics[width=1.\textwidth]{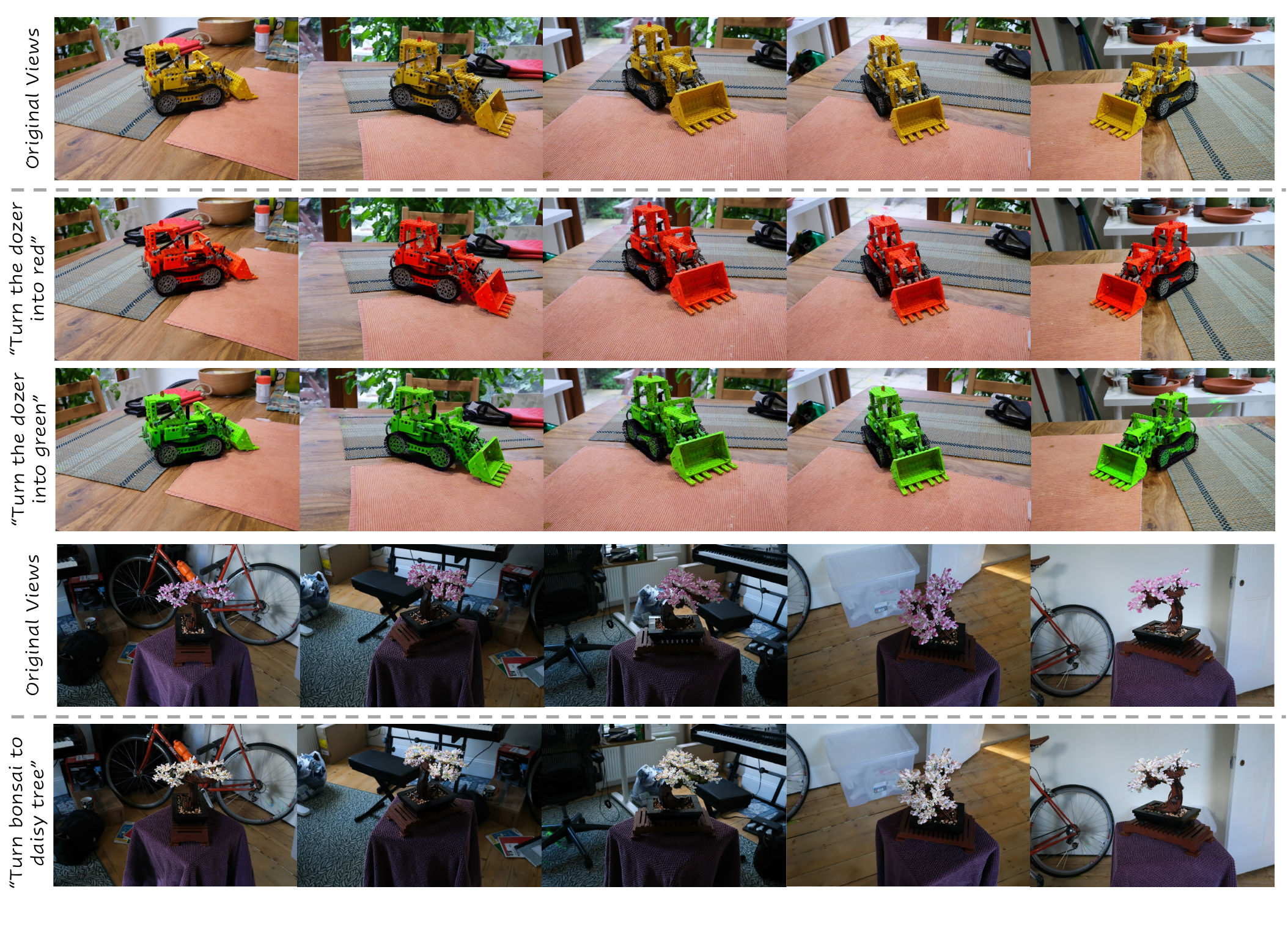}
    \caption{Illustrative visual examples on ``kitchen'' and ``bonsai'' scenes from Mip-Nerf 360~\cite{barron2022mip} dataset with various text prompts  showcasing the application of our proposed method for local editing on a 3D scene, demonstrating its effectiveness in handling fine-grained modifications.
}
    \label{fig:res_dozer}
\end{figure*}

\subsection{Implementation Details}
The proposed method is implemented using PyTorch and trained on one NVIDIA A6000 GPUs.
We use the optimized renderer from~\cite{kerbl20233d} for Gaussian rendering, with our implementation based on Threestudio~\cite{guo2023threestudio}. All 3D Gaussians in this study are trained using the method described in~\cite{kerbl20233d}. For 2D editing, we employ IP2P~\cite{brooks2023instructpix2pix} following most prior methods~\cite{chen2024dge,chen2024gaussianeditor,haque2023instruct} to ensure fair comparison. A selection of 3D scenes from publicly available datasets, such as LLFF and Mip-NeRF360~\cite{barron2022mip,mildenhall2019local}, is used for evaluation.The 3DGS model is finetuned for 1,000-1,500 iterations, with a default classifier-free guidance strength of 7.5. For certain cases, we adjust the edit strength consistently across all competing methods. 

\vspace{1mm}
\noindent\textbf{Training details.}
For the training iterations, the 3DGS model is fine-tuned for 1,000 iterations on the ``kitchen'' scene, as shown in Figure~\ref{fig:res_dozer}, and for 1,500 iterations on other scenes. 
For every 500 iterations, we repeat the cycle mentioned in Section~4.3 of main paper and re-select a new frontal view.
We select a linearly decreasing starting timestep for IP2P, as high-level semantics are adjusted after some iterations, leaving only low-level structural details to be refined in subsequent cycles. Specifically, \( t = [750, 500, 250] \) is used as the starting denoising steps for three training cycles, and \( t = [750, 500] \) is used for two training cycles.
For each cycle, we select 20 adjacent views for 2D editing.
Besides, a default classifier-free guidance strength of 7.5 is used, except for the ``face'' scene in Figure~1 of main paper with prompts such as ``make him older'' and ``turn him into a werewolf.'' A classifier-free guidance strength of 7.5 effectively handles most attribute and accessory editing cases.
For the localization part, we set \(\tau = 600\) steps and the mask threshold \(\gamma = 0.6\) for most scenes.

\subsection{Baselines and Evaluation Metrics}
We compare our approach with a set of competing methods, including two NeRF-based editing techniques—Instruct-N2N~\cite{haque2023instruct} and ViCA-NeRF~\cite{dong2024vica}—and three 3D Gaussian Splatting (3DGS)-based methods: GaussianEditor~\cite{chen2024gaussianeditor}, DGE~\cite{chen2024dge}, and Gaussctrl~\cite{gaussctrl2024}. All methods, including ours, utilize IP2P as the 2D editor, except for Gaussctrl, which uses ControlNet~\cite{zhang2023adding}.
To evaluate these methods, we use CLIP Text-Image Direction Similarity (CTIDS)~\cite{radford2021learning} to assess semantic alignment after editing, along with Image-Image Similarity (IIS) and FID~\cite{heusel2017gans} to measure fidelity. We also measure editing time, following the evaluation metrics used in~\cite{wang2024gaussianeditor}.
Furthermore, we conduct a user study to assess subjective aspects, specifically measuring the detailed structural quality and consistency across different views, results and details of which are included in the supplementary.

\begin{figure}[!t]
    \centering
    \includegraphics[width=.95\linewidth]{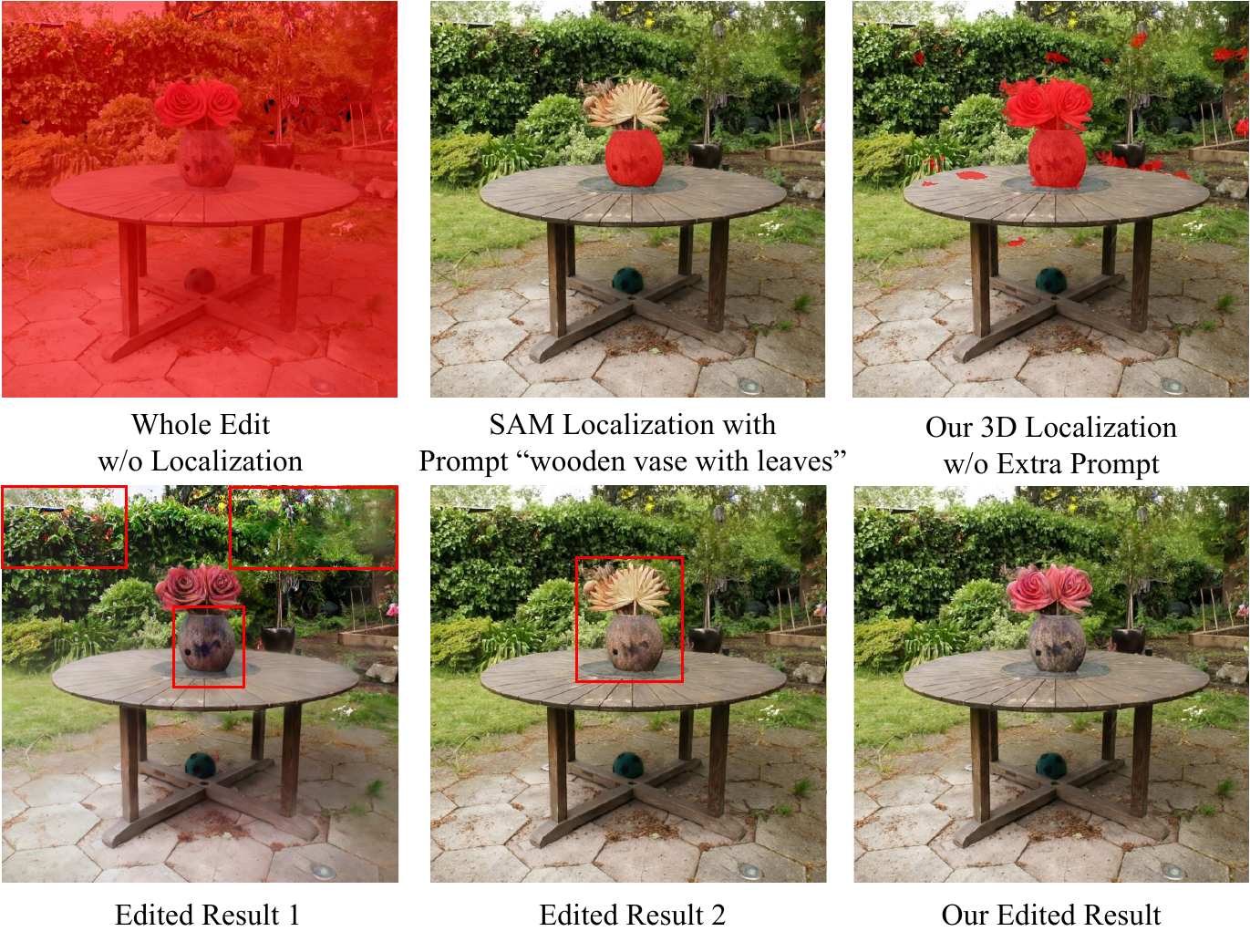}
    \caption{Visualization of our 3D localization and the corresponding edited result compared to the whole region editing and directly using SAM with extra prompt for localization.
    }
    \label{fig:local}
\end{figure}

\begin{figure}[t]
    \centering
    \includegraphics[width=1.\linewidth]{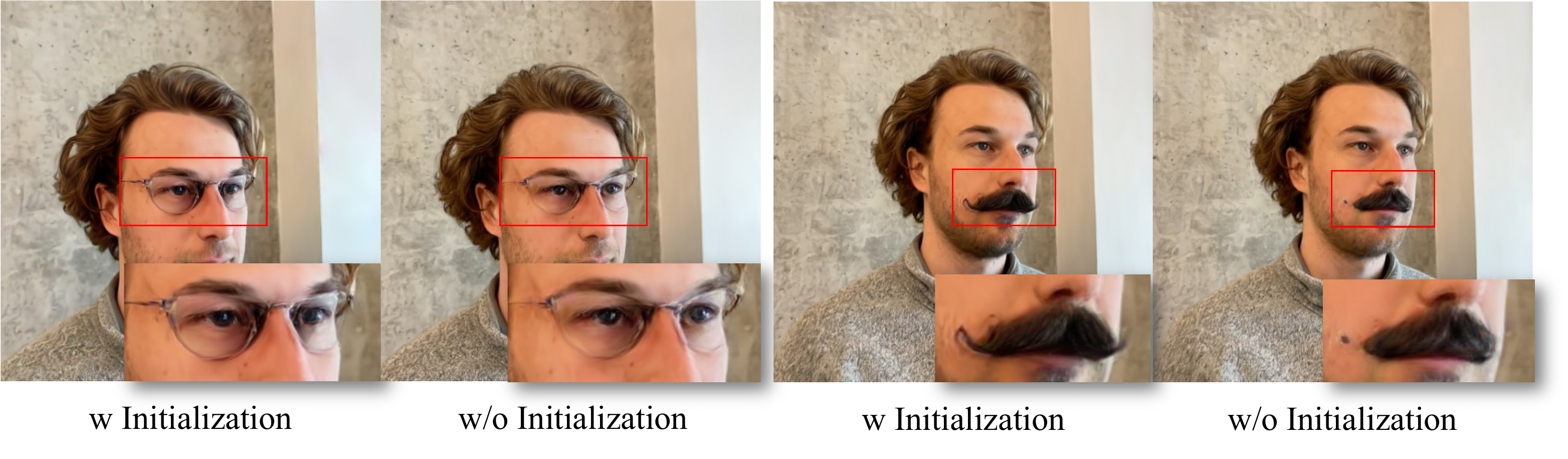}
    \caption{Two editing examples of our method with and without initialization. Without proper initialization, if parts of these accessories aren't set to slightly extend beyond the character's edges, they may be omitted in certain views, resulting in inconsistencies and undesired outcomes.
    }
    \label{fig:initial}
\end{figure}

\begin{figure}[t]
    \centering
    \includegraphics[width=1.\linewidth]{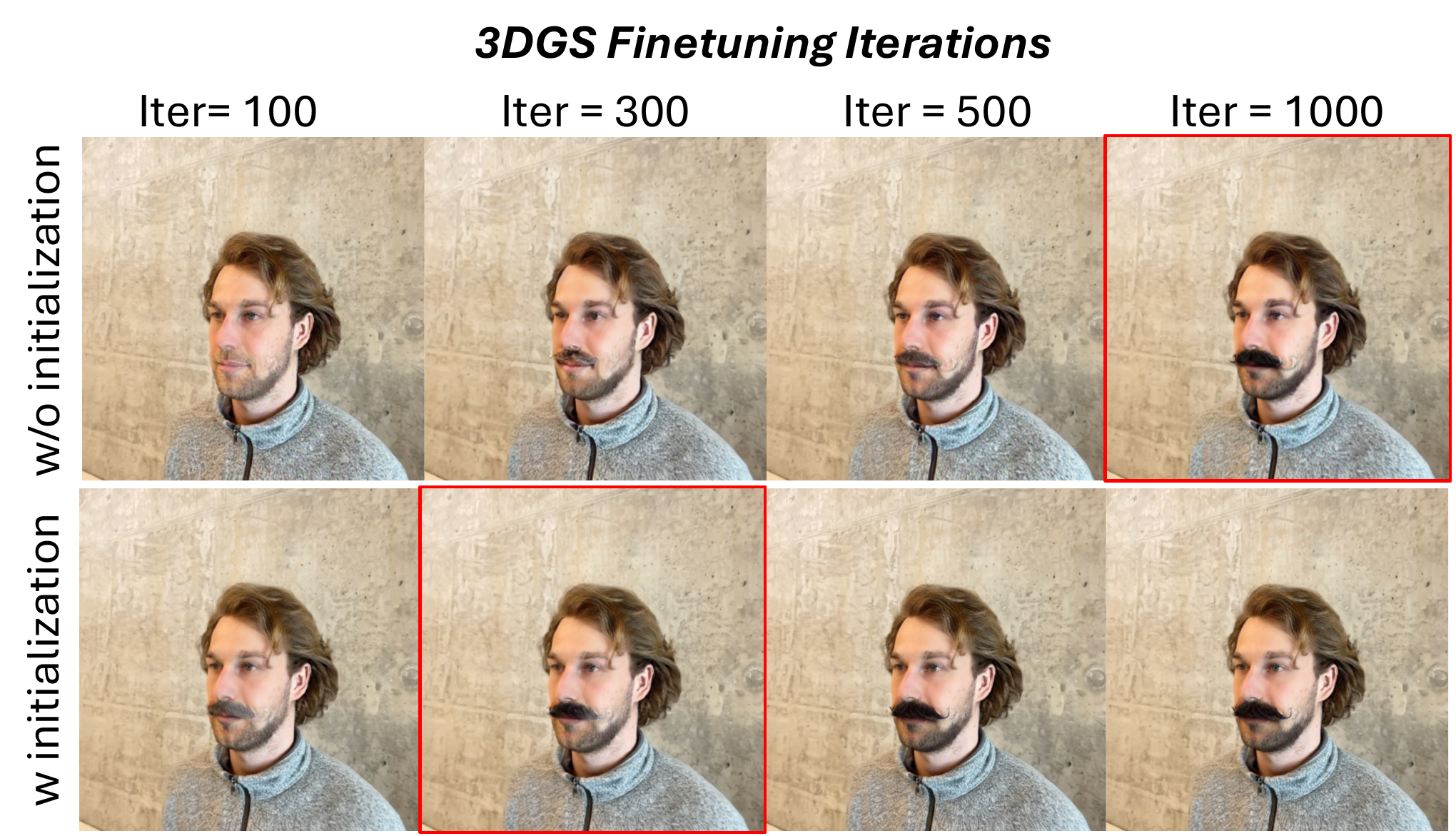}
    \caption{Visualization of intermediate results during 3DGS finetuning with and without initialization. Proper initialization effectively reduces the required finetuning iterations.
    }
    \label{fig:visual}
\end{figure}

\begin{figure*}[h]
    \centering
    \includegraphics[width=1.\textwidth]{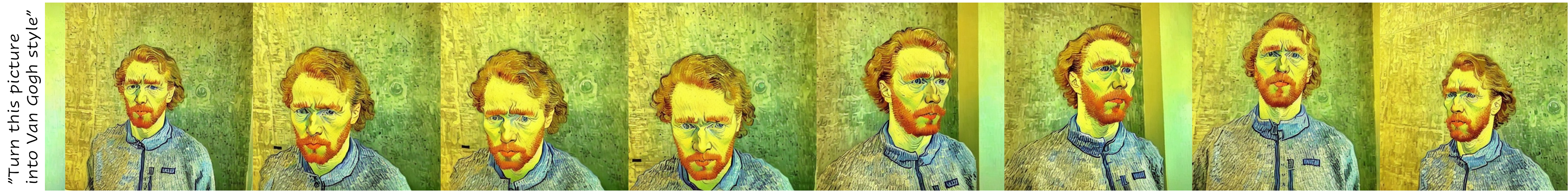}
    \caption{An example of global editing is setting $\gamma=0$ for an all-one 3D localization.
}
    \label{fig:global}
\end{figure*}


\subsection{Qualitative Evaluation}
Figure~\ref{fig:res} shows a comparison between our method and competing methods across different views on two scenes, each accompanied by the corresponding text prompt.
GaussianCtrl~\cite{gaussctrl2024} utilizes ControlNet as the 2D editor, which is more suitable for global editing, offering better consistency and aesthetic quality. However, it struggles to retain original scene information, often altering features such as face ID and skin tone during editing. Noted that we followed the GaussianCtrl using pre-processing data to a $512^2$ resolution, and we manually padded the visual examples in Figure~\ref{fig:res} for better visualization.
DGE~\cite{chen2024dge} and GaussianEditor~\cite{chen2024gaussianeditor} can maintain high-level information consistency; however, artifacts may appear in boundary details, such as around the edges of an added mustache. Additionally, fine-grained elements like clothing tags and zippers may lose some high-frequency details, resulting in slight distortions.
Apart from the issue of detail preservation, the second example highlights an information leakage problem. Existing methods may inadvertently modify unintended regions that are closely related to the text prompt, such as background elements like grass and leaves. This can lead to inconsistencies across views and the appearance of unwanted artifacts.
In contrast, towards local editing tasks, our proposed method demonstrates the ability to achieve enhanced fine-grained structural details, particularly evident in the reduced occurrence of artifacts along the editing boundaries. 
Moreover, our method excels in maintaining high fidelity to the original scene, effectively preserving key identity-related attributes, such as face ID, clothing details, and hair style after editing. 
We include additional qualitative results in Figure~\ref{fig:res_combined}, showcasing multiple edited views. The results demonstrate the effectiveness of our method, exhibiting strong consistency across different viewpoints.
We primarily focus on editing attributes such as color, texture, and facial features, as well as accessories like hats, clothing, and pants.
For local editing tasks, our proposed method showcases its capability to achieve refined fine-grained structural details, with a notable reduction in artifacts along editing boundaries. Additionally, it demonstrates strong fidelity to the original scene by effectively preserving key identity-related attributes such as facial features, clothing details, and hairstyle after editing.


\subsection{Quantitative Evaluation}
Following previous works~\cite{haque2023instruct,chen2024dge}, we evaluate the alignment of edited 3D models and target text prompts with CLIP similarity score, \textit{i.e.}, the cosine similarity between the text and image embeddings encoded by CLIP.
We also evaluate the fidelity preservation between original and edited 3D models by calculating the cosine similarity between original image and edited image as well as the FID between original and edited image sets.
Table~\ref{tab:my_label} presents a detailed comparison of our method against several competing approaches including editing results on three 3D scenes, where the details can be found in supplementary. Our method demonstrates comparable CLIP Text-Image Direction Similarity scores to the current state-of-the-art methods, indicating its effectiveness in preserving textual alignment. Additionally, our approach outperforms competing methods in preserving directional consistency with the original scene, ensuring high-fidelity editing that selectively modifies only the intended characteristics of the scene.
Apart from evaluating effectiveness, we also assess the efficiency of our approach. 
By leveraging coarse initialization in Section~\ref{sec:init} and skipping diffusion steps during iterative refinement in Section~\ref{sec:refine}, our method achieves more efficient editing, offering up to a 4$\times$ speedup compared to existing 3DGS-based methods and a 25$\times$ speedup compared to NeRF-based methods.


\subsection{User Study}
We conducted a user study focusing on two key aspects: view consistency and fine-grained details. Our proposed method was compared against GaussCtrl~\cite{gaussctrl2024}, DGE~\cite{chen2024dge}, and GaussianEditor~\cite{chen2024gaussianeditor} on the several scenes, with the participation of 30 users.
Our proposed method also achieves a voting percentage of 61.67\%, with GaussCtrl, DGE, and GaussianEditor receiving 8.33\%, 18.33\%, and 11.67\%, respectively.

\subsection{Ablation Study}

\noindent\textbf{The effect of 3D localization.}
To validate the effect of the 3D localization, we conduct experiments on a variant where the 3D localization component is removed from our framework. As shown in the comparison between the first column and third column in Figure~\ref{fig:local}, removing the 3D localization results in distortion and blurry in background and unedited regions, 
a significant decrease in the overall scene reconstructed quality, especially in scenarios with complex spatial configurations. 
Color information can leak into unrelated or undesired areas, leading to unintended color changes in objects like the vase, table, and floor. For instance, they might turn somewhat red due to the influence of a prompt focused on ``rose''.
Besides, our 3D localization is distinct from most previous works~\cite{chen2024gaussianeditor,chen2024dge,wang2025view} used SAM with a corresponding prompt to locate the edited regions in object level.
As shown in the comparison between the second and third column in Figure~\ref{fig:local}, some editing scenarios make it challenging to accurately target a specific object, such as object replacement or adding extra accessories. Sometimes, SAM may fail even if you provide a fairly precise prompt.
In contrast, our 3D localization identifies the edited regions without requiring additional prompts and more effectively prevents information leakage into unedited areas.

\vspace{1mm}
\noindent\textbf{The effect of initialization by predicted depth.}
We utilize the edited first view as a reference, generating its depth map using the DepthAnything model. This depth map is then inversely rendered into 3D space to produce a coarse point cloud for initialization. This initial coarse representation effectively reduces the overall training time and enables the edited shape and accessories in the first view to be consistently rendered across other views, serving as a condition for subsequent view edits.
As illustrated in Figure~\ref{fig:initial}, without a robust initialization, adding new accessories or altering the shape of an existing object may encounter significant challenges in localization. This is especially true for details like the edges of accessories, which are difficult to generate if they extend beyond the object’s boundary in certain views. For example, in our case with glasses and a mustache, if parts of these accessories are not well-initialized to extend slightly beyond the character’s edges, they may be omitted in certain views, leading to inconsistencies. Thus, a good initialization serves as a crucial solution to mitigate these localization limitations and ensure cohesive multi-view rendering.

\subsection{Network Analysis}

\vspace{1mm}
\noindent\textbf{Efficiency analysis.} 
With the proposed 3D localization and regional initialization, both the diffusion process and 3DGS fine-tuning are accelerated.
On one hand, for IP2P sampling steps, we employ a linearly decreasing starting timestep, as high-level semantics are adjusted in the early iterations, leaving only low-level structural details to be refined in later stages. This approach effectively reduces the diffusion refinement time. (Details refer to supplementary.)
On the other hand, since the coarse shape of the edited region (e.g., the mustache) is well-initialized, the necessary Gaussian points are already properly positioned. Only specific properties, such as texture and color, require further refinement using the diffusion model. As a result, the number of iterations for 3DGS fine-tuning is significantly reduced.
As shown in Figure~\ref{fig:visual}, with proper initialization, noticeable modifications occur after just 300 iterations, whereas without initialization, up to 1,000 iterations may be required.

\vspace{1mm}
\noindent\textbf{Global editing.} 
Our method focuses on viewpoint consistency and high fidelity in localized editing, making it particularly effective for attribute modifications. However, for global edits like style transfer, 3D localization can be treated as an all-one mask. In this case, the required 3DGS finetuning iterations increase, as the initialized 3DGS may lack the precision of localized editing, necessitating further refinement to achieve the desired outcome. 
Figure~\ref{fig:global} shows a style transfer example on the ``face'' scene, requiring 2,000 3DGS finetuning iterations over four cycles.

%% file: sec/4_concl.tex
\vspace{1mm}
\noindent\textbf{Limitations.}
Our method relies on an additional depth estimation module to provide an initial 3D reconstruction of the scene, which serves as the starting point for the subsequent optimization process. This initialization is critical to guiding the editing process toward geometrically consistent and view-aligned results. However, in scenarios where the depth estimation is highly inaccurate, such as under extreme lighting conditions, low-texture regions, reflective surfaces, or occlusions, the initialization quality deteriorates significantly. As a result, the optimization is more prone to start from an incorrect geometry and requires substantially more iterations to converge to a satisfactory solution. The model must effectively compensate for these inaccuracies through iterative refinement, gradually correcting the structural inconsistencies and recovering a faithful reconstruction. While our pipeline is still capable of handling such challenging cases, the overall computational cost increases and the editing fidelity may initially be compromised due to the suboptimal guidance from the erroneous depth.


\section{Conclusion}
In conclusion, our proposed framework for 3D Gaussian Splatting (3DGS) offers a highly efficient and effective solution for localized editing in 3D scenes. By integrating 2D diffusion-based editing to identify and guide region-specific modifications across multiple views, and leveraging depth-based initialization derived from a pretrained 2D foundation model, our method significantly improves both the precision and coherence of edits in the 3D space. This design enables an iterative, view-consistent refinement process that preserves structural integrity and textural fidelity across viewpoints. Extensive experiments demonstrate that our approach not only achieves state-of-the-art editing quality but also delivers a substantial speedup—up to $4\times$ faster—compared to existing 3DGS-based editing pipelines, making it a practical and scalable choice for real-world applications.